\documentclass[review]{elsarticle}

\usepackage{lineno,hyperref}
\usepackage{CJK}
\usepackage{array}
\usepackage{amsmath,amsfonts,amssymb,graphicx}
\usepackage{cases}
\usepackage{graphicx}
\usepackage{caption}
\usepackage{graphicx}
\usepackage{picinpar}
\usepackage{multirow}
\usepackage{lineno,hyperref}
\usepackage{setspace}

\usepackage{algorithm}
\usepackage{algorithmic}
\usepackage{amsmath,amssymb,amsfonts, bm}

\modulolinenumbers[5]

\journal{}

\begin{document}

\begin{frontmatter}

\title{Multi-view Locality Low-rank Embedding for Dimension Reduction}

\author[mymainaddress]{Lin Feng\corref{mycorrespondingauthor}}
\cortext[mycorrespondingauthor]{Corresponding author}
\ead{fenglin@dlut.edu.cn}

\author[mymainaddress]{Xiangzhu Meng}
\author[mysecondaryaddress]{Huibing Wang}

\address[mymainaddress]{School of Computer Science and Technology, Dalian University of Technology, Dalian, 116024, PR China}
\address[mysecondaryaddress]{Information Science and Technology College, Dalian Maritime University, Dalian 116024, PR China}

\begin{abstract}
During the last decades, we have witnessed a surge of interests of learning a low-dimensional space with discriminative information from one single view. Even though most of them can achieve satisfactory performance in some certain situations, they fail to fully consider the information from multiple views which are highly relevant but sometimes look different from each other. Besides, correlations between features from multiple views always vary greatly, which challenges multi-view subspace learning. Therefore, how to learn an appropriate subspace which can maintain valuable information from multi-view features is of vital importance but challenging. To tackle this problem, this paper proposes a novel multi-view dimension reduction method named Multi-view Locality Low-rank Embedding for Dimension Reduction ($MvL^2E$). $MvL^2E$ makes full use of correlations between multi-view features by adopting low-rank representations. Meanwhile, it aims to maintain the correlations and construct a suitable manifold space to capture the low-dimensional embedding for multi-view features. A centroid based scheme is designed to force multiple views to learn from each other. And an iterative alternating strategy is developed to obtain the optimal solution of $MvL^2E$. The proposed method is evaluated on 5 benchmark datasets. Comprehensive experiments show that our proposed $MvL^2E$ can achieve comparable performance with previous approaches proposed in recent literatures.
\end{abstract}

\begin{keyword}
Multi-view learning \sep Low rank \sep Dimension reduction
\end{keyword}

\end{frontmatter}

\section{Introduction}

\begin{figure}[!htb]
\centering
\includegraphics[width=5in]{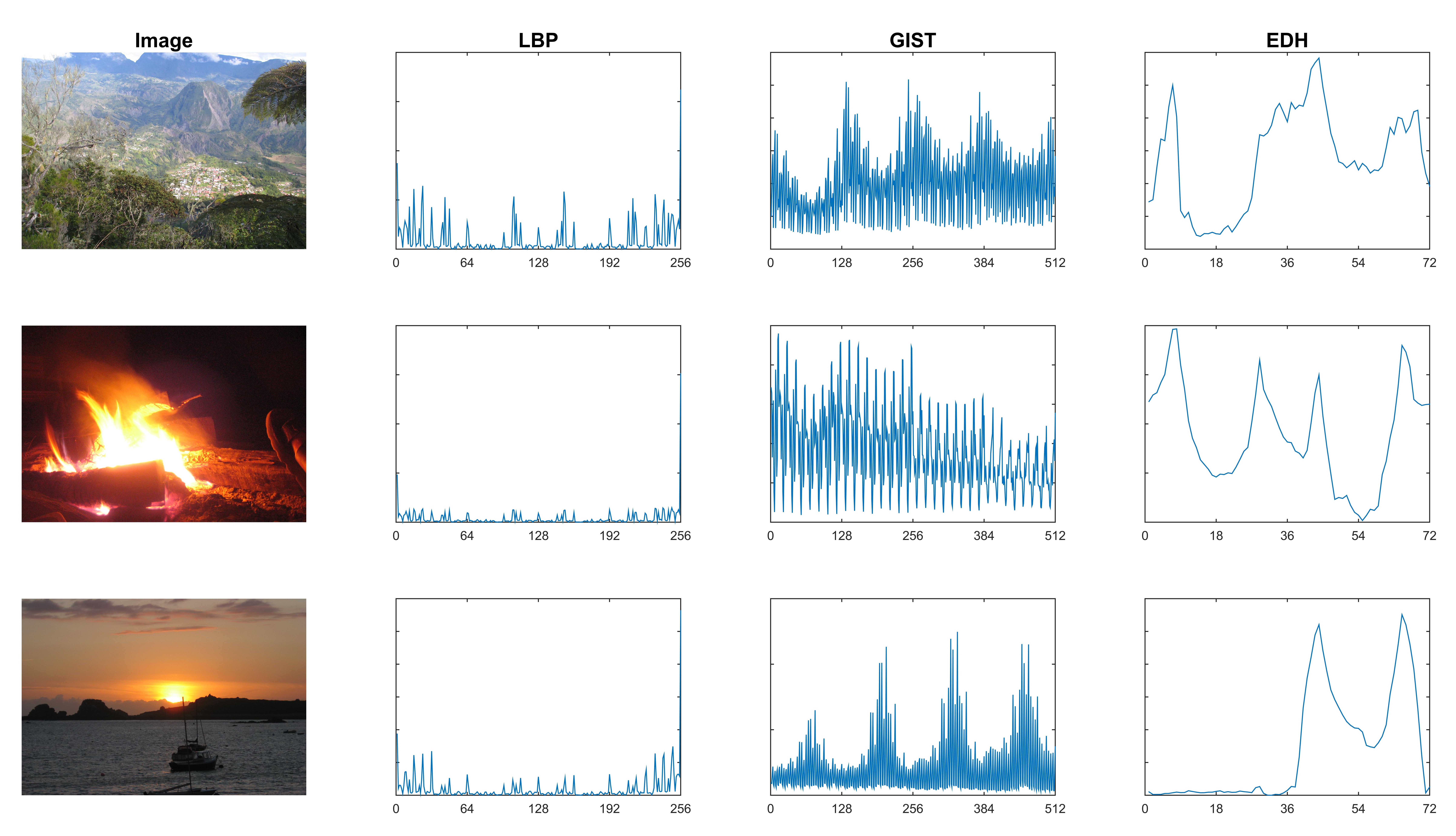}
\caption{Multi-view features in representing the images}
\label{Fig1}
\end{figure}

In many real world applications, one object can always be characterized at different kinds of viewpoints\cite{li2018survey}\cite{wang2012multimodal}\cite{xu2013survey}. For examples, webpages usually consist of both the page-text and hyperlink information; An image could be described with color, text or shape information, such as HSV, Local Binary Pattern (LBP)\cite{ojala2002multiresolution}, Gist\cite{douze2009evaluation}, Histogram of Gradients (HoG)\cite{dalal2005histograms}, Edge Direction Histogram (EDH)\cite{gao2008image}(as Fig.\ref{Fig1}). Therefore, multiple views contain more useful information than just one single view. It can improve the performance of most applications by making full use of the complementary information from multiple views. However, most extracted features in many applications usually locate in high-dimensional spaces, such as text classification\cite{jiang2011fuzzy,wang2015visual}, face recognition\cite{tao2012discriminative}\cite{qiao2010sparsity} and image retrieval\cite{wang2012view}, \cite{wang2012event}. Due to the huge time consumption and computation cost on directly processing these high-dimensional features, a variety of dimensional reduction methods are proposed to tackle this problem. They learn a low-dimensional subspace by preserving enough semantic information of the samples. Principle Components Analysis (PCA)\cite{wold1987principal} and Linear Discriminant Analysis (LDA)\cite{mika1999fisher} are two popular linear DR methods which fully maintain the global Euclidean structure of multi-view features. PCA is an unsupervised DR method which captures the low-dimensional subspace by maximizing the variances of samples. Contrast to PCA, LDA is a supervised DR method to maximize the ratio between the trace of between-class scatter and the trace of within-class scatter. Besides the investigations for the global structure in samples, the local correlation between samples is worthy of attention. Many DR methods that attempt to apply local correlations have been proposed in the past decades, such as Locality Preserving Projections (LPP)\cite{he2004locality}, Neighborhood Preserving Embedding (NPE)\cite{he2005neighborhood}, and Locality Sensitive Discriminant Analysis (LSDA)\cite{cai2007locality}. Unlike these linear methods above, varieties of manifold learning methods have been proposed to deal with the nonlinear high-dimensional feature, which lies on a sub-manifold of the observations space, such as Isometric Mapping (Isomap)\cite{tenenbaum2000global}, Laplacian Embedding (LE)\cite{belkin2003laplacian} and Local Linear Embedding (LLE)\cite{roweis2000nonlinear}. Besides, low rank normalization has been gained much attention in recent years. For example, robust PCA is presented in \cite{xu2010robust} to recover the correct column space of the uncorrupted matrix by involving matrix decomposition using nuclear norm minimization, and the work \cite{liu2013robust} proposes a Low-Rank Representation (LRR) method, which seeks the lowest rank representation among all the candidates that can represent the data samples as linear combinations of the bases in a given dictionary. However, these DR methods mainly focus on single view features, and couldn't be directly applied to multi-view cases due to information integration with compatibility and complementary of multi-view features.

On integrating rich information among different features, much progress of multi-view learning methods has been made in developing effective multi-view methods. The work\cite{chaudhuri2009multi} proposes that Canonical Correlation Analysis (CCA)\cite{hardoon2004canonical} could be used to project the two view into the common subspace by maximizing the cross correlation between two views. Furthermore, CCA is further generalized for multi-view scenario termed as multi-view canonical correlation analysis (MCCA)\cite{rupnik2010multi}. Multi-View Discriminant Analysis\cite{kan2016multi} is proposed to extend LDA into a multi-view setting, which projects multi-view features to one discriminative common subspace. The paper\cite{zhang2018generalized} proposes a Generalized Latent Multi-View Subspace Clustering, which jointly learns the latent representation and multi-view subspace representation within the unified framework. Besides these multi-view learning methods, some researches based on multiple graph learning have been developed. Multiview Spectral Embedding (MSE)\cite{xia2010multiview} incorporates conventional algorithms with multiview data to find a common low-dimensional subspace, which exploits low-dimensional representations based on graph. Meanwhile, it's attracted wide attention to achieve the multi-view clustering agreement \cite{kumar2011co}, \cite{wang2015robust}, \cite{wang2016iterative}, \cite{wang2018multiview} to yield a substantial superior clustering performance over the single view paradigm. For example, the work \cite{kumar2011co} aims to propose a co-regularized multi-view spectral clustering framework that captures complementary information among different viewpoints by co-regularizing a clustering hypotheses. Besides, such works in \cite{wu20193}\cite{wang2017effective}\cite{wu2019cycle}\cite{wang2018beyond} also obtain promising performance in multi-view learning environment.

\subsection{Contributions}
In this paper, we first propose a new DR method to maintain the low rank local linear structure in the geometric manifold space, called Locality Low-rank Embedding ($L^2E$). Then, we extend the $L^2E$ for the single view to the multi-view framework to propose a multi-view DR method based on the centroid manifold structure called Multi-view Locality Low-rank Embedding for Dimension Reduction ($MvL^2E$), which aims to find a suitable manifold space to capture the low-dimensional embedding from multi-view data while maintains low rank local manifold structure from each view. It's a multi-view scheme designed to integrate multi-view features to one common manifold space. Then, an optimization algorithm using iterative alternating strategy is developed to obtain the optimal solution of $MvL^2E$. The contributions in this paper are illustrated as follows. First, a new DR method called $L^2E$ is proposed and then extended into the multi-view framework to seek a low-dimensional embedding in a common manifold space. Second, we propose an effective and robust iterative method to seek an optimal solution for $MvL^2E$. Third, not only the low-dimensional feature in a common manifold space has reliable performance, but also the single feature corrected and complemented by ones from the others views obtains more outstanding performance than raw single view features.

\subsection{Organization}
The rest of the paper is organized as follows. In Section 2, we provide briefly some related methods which have attracted extensive attention. In Section 3, we describe the construction procedure of $MvL^2E$ and optimization algorithm for $MvL^2E$ in detail. In Section 4, empirical evaluations based on the applications of text classification and image classification demonstrate the effectiveness of our proposed approach. In Section 5, we make a conclusion of this paper.

\section{Related Works}
In this section, we first introduce a classical spectral embedding DR method which learns the cluster memberships information by exploiting the graph Laplacian matrix. Then, we review a multi-view clustering method, which is a method based on agreement called co-regularized multi-view spectral clustering\cite{kumar2011co}.

\subsection{Spectral Embedding}
Let $\bm{X}=\left\{ {\bm{x}_1,\bm{x}_{_2}, \ldots ,\bm{x}_N} \right\}$ denote the features set and $\bm{K}$ denote the similarity matrix of the features set $\bm{X}$. In addition, let $\bm{D}$ denote a diagonal matrix and ${\bm{D}_{ii}}=\sum\limits_{j = 1}^N {{\bm{K}_{ij}}}$. We write the normalized graph Laplacian matrix for the view as $\bm{L}=\bm{D}^{^{ - 1/2}}\bm{K}\bm{D}^{^{ - 1/2}}$. The spectral clustering method\cite{ng2002spectral} solves the following optimization problem for the normalized graph Laplacian matrix $\bm{L}$:

\begin{equation}.
    \begin{array}{l}
\mathop {\max }\limits_{\bm{U}} \hspace{0.5em}tr(\bm{U}^{^T}\bm{L}\bm{U})\\
\hspace{0.5em}s.t.\hspace{1em}\bm{U}^{^T}\bm{U} = \bm{I}
\end{array}
\end{equation}

The rows of matrix $\bm{U}$ are the embedding of the features set that could be given to the k-means algorithm to obtain cluster memberships.

\subsection{Co-regularized Multi-view Spectral Clustering}

Co-regularized Multi-view Spectral Clustering\cite{kumar2011co} is a spectral clustering algorithm that achieves this goal by co-regularizing the clustering hypotheses across views. Assume that given data has multiple views. Let ${\bm{X}^v}=\left\{ {\bm{x}_1^v,\bm{x}_{_{}^2}^v, \ldots ,\bm{x}_N^v} \right\}$ denote the features set in the $v$th view and $\bm{K}^v$ denote the similarity matrix of $\bm{X}^v$ in this view. We write the normalized graph Laplacian matrix for the view as: ${\bm{L}^v}={\bm{D}^v}^{^{ - 1/2}}{\bm{K}^v}{\bm{D}^v}^{^{ - 1/2}}$, where $\bm{D}^v$ is a diagonal matrix and $\bm{D}_{ii}^v= \sum\limits_{j = 1}^N {\bm{K}_{ij}^v}$. The method of Co-regularized Multi-view Spectral Clustering builds on the standard spectral clustering above by appealing to the co-regularized framework, which makes the clustering hypotheses on different views agree with each other. The cost function for the measure of disagreement between clusters of the view $\bm{X}^v$ and the view $\bm{X}^w$ can be defined as follows:
\begin{equation}\label{co_regularizaton}
    D\left( {{\bm{U}^v},{\bm{U}^w}} \right) =  - tr\left( {{\bm{U}^v}{\bm{U}^v}^{^T}{\bm{U}^w}{\bm{U}^w}^{^T}} \right)
\end{equation}
where the matrix $\bm{U}^v$ and the matrix $\bm{U}^w$ represent the embedding of the view $\bm{X}^v$ and the view $\bm{X}^w$ respectively. Therefore, combining Eq.(\ref{co_regularizaton}) with the spectral clustering objectives of the single view, we could get the following joint maximization problem for $m$ views:
\begin{equation}
\begin{split}
&\mathop {\max }\limits_{{\bm{U}^1},{\bm{U}^2}, \ldots ,{\bm{U}^m} \in {\mathbb{R}^{N \times k}}} \sum\limits_{v = 1}^m {tr({\bm{U}^v}^{^T}{\bm{L}^v}{\bm{U}^v})}  + \vspace{1cm}\\
&\hspace{2em}\lambda \sum\limits_{1 \le v,w \le m} {tr\left( {{\bm{U}^v}{\bm{U}^v}^{^T}{\bm{U}^w}{\bm{U}^w}^{^T}} \right)} \\
&\hspace{3em}s.t.\hspace{1.5em}{\bm{U}^v}^{^T}{\bm{U}^v}{ =  I,} \forall {\rm{  1}} \le v \le m{\rm{    }}    \\
\end{split}
\end{equation}
where the hyperparameter $\lambda$ trades-off the spectral clustering objectives and the spectral embedding disagreement term. To solve the loss objective problem, the iterative alternating strategy could be employed.

\section{Multi-view Locality Low-rank Embedding for Dimension Reduction}
In this section, We first propose a new DR method called Locality Low-rank Embedding ($L^2E$) which maintains the low rank local linear structure in the geometric manifold space in Section \ref{single_view_$L^2E$}. Then, we extend the $L^2E$ for the single view into the multi-view framework to propose a method called Multi-view Locality Low-rank Embedding for Dimension Reduction ($MvL^2E$) in Section \ref{construction}, which fully integrates compatible and complementary information from multi-view features sets to construct low-dimensional embedding. Finally, an iterative alternating strategy is adopted to find the optimal solution of $MvL^2E$ and the optimization procedure of $MvL^2E$ is illustrated in detail in Section \ref{optimization}.

\subsection{Locality Low-rank Embedding}\label{single_view_$L^2E$}
Low Rank Representation based method is quite superior in terms of its effectiveness, intuitiveness and robustness to noise corruptions, which deals with subspace recovery problem. Assume that we are provided a features set consisting of $N$ samples, which are extracted from the $v$th view. We express the features set in the $v$th view  as $\bm{X}^v=[{\bm{x}_1^v}, {\bm{x}_2^v},  \ldots , {\bm{x}_N^v}] \in {\mathbb{R}^{D^v \times N}}$, where $\bm{D}^v$ is the dimension of features set. When we choose the matrix $\bm{X}^v$ itself as a dictionary that linearly spans the data space. We could get the following optimization problem:
 \begin{equation}
 \begin{split}
 &\mathop {\min }\limits_{\bm{Z}^v,\bm{E}^v} { rank(\bm{Z}^{v}) + }\lambda {\left\| \bm{E}^v \right\|_{2,1}} \\
 & s.t.\hspace{0.5em}{ \bm{X}^{v} = \bm{X}^{v}\bm{Z}^{v} + \bm{E}^{v}}\\
 \end{split}
 \end{equation}
where $\lambda$ is a hyperparameter and $\bm{Z}^v \in {\mathbb{R}^{N \times N}}$ is the lowest rank representation of data $\bm{X}^v$. Even though this method makes full use of this hypothesis that the data is considered as samples approximately drawn from a mixture of several low-rank subspaces, local structure information in samples space could be more able to reflect the relations among samples beyond global structure. To further investigate local structure with low rank representation, we choose the dynamic dictionary for individual sample by using $K$ its near neighbors. To combine the low rank hypothesis, we could get  the  following optimization problem:
 \begin{equation}
 \begin{split}
 & \mathop {\min }\limits_{\bm{Z}^v,\bm{E}^v} {rank(\bm{Z}^v) + }\lambda {\left\| \bm{E}^v \right\|_F^2} \\
 & s.t.\hspace{0.5em}{\bm{X}_i^v = {\bm{D}_i^v\bm{Z}_i^v} + {\bm{E}_i^v}, {\forall 1 \le i \le N}} \\
 \end{split}
 \end{equation}
where ${\bm{D}_i^v} \in {\mathbb{R}^{D^v \times K}}$ is the dictionary of $i$th sample consisting of $K$ its closed neighbors and $\bm{Z}_i^v$ and $\bm{E}_i^v$ denote the $i$th column data in the matrix $\bm{Z}^v$ and $\bm{E}^v$ respectively. It is easy to see that the solution to the above equation may not be unique. As a common practice in rank minimization problems, we replace the rank function with the nuclear norm and subject to the constraints the columns of the weight sum to one, resulting in the following optimization problem:
  \begin{equation}\label{$L^2E$}
  \begin{split}
  &\mathop {\min }\limits_{\bm{Z}^v,\bm{E}^v} {\| \bm{Z}^v \|_* + }\lambda {\| \bm{E}^v \|_F^2} \\
  &{ s}{.t.}\hspace{0.5em}{ \bm{X}_i^v = \bm{D}_i^v\bm{Z}_i^v + \bm{E}_i^v, {\bm{Z}_i^v}^T\bm{1}=1, \forall 1 \le i \le N}\\
 \end{split}
 \end{equation}

To solve this equation, we propose a two-stage approach. First, we fix the first term in the objective function to exchange the problem into optimization problem with closed solution $\bm{Z}^v$. Second, we apply the Singular Value Thresholding (SVT)\cite{cai2010singular} operator to solve the low rank solution according to $\bm{Z}^v$ solved in previous step. However, the solved low rank representation couldn't be directly used as discriminative foundation in applications because base dictionary based on near neighbors is different from each other.

It's not difficult to discover the conclusion that the matrix $\bm{Z}^v$ reflects the local linear structure of samples. To make use of the matrix $\bm{Z}^v$, we transform the raw features into a lower dimensionality space while maintain the low rank linear structure. We suppose that the data lie on or near a smooth nonlinear manifold of lower dimensionality $d^v \ll D^v$. The low rank weight matrix $\bm{Z}^v$ reflects intrinsic geometric properties of the features set. And we expect their characterization of local geometry in the original data space to be equally valid for local patches on the manifold. Therefore, the low rank coefficients $\bm{Z}_i^v$ that reconstruct the $i$th data point in $\bm{D}^v$ dimensions should also reconstruct its embedded manifold coordinates in $d^v$ dimensions. According to the idea above, each high-dimensional feature $\bm{x}_i^v$ is mapped to a low-dimensional embedding $\bm{y}_i^v$ representing global internal coordinates on the manifold. For the convenience of modeling and solving, a simple trick is used to transform the matrix $\bm{Z}^v \in {\mathbb{R}^{K \times N}}$ into a matrix $\bm{W}^v \in {\mathbb{R}^{N \times N}}$, which fills column elements in the matrix $\bm{W}^v$ according to the low rank coefficients $\bm{Z}_i^v$ of its neighbors and fills zeros into other elements. To solve all low-dimensional embedding $\bm{Y}^v=\left[ {{\bm{y}_1^v},{\bm{y}_2^v}, \ldots ,{\bm{y}_N^v}} \right] \in {\mathbb{R}^{d^v \times N}}$, we minimize the embedding cost function

\begin{equation}
    \Phi \left( \bm{Y}^v \right){=}{\sum\limits_{i = 1}^N {\| {{\bm{y}_i^v} - \sum\limits_j {{\bm{W}_{ij}^v}{\bm{y}_j^v}} } \|_2 ^2}}
\end{equation}

To avoid degenerate solutions, we constrain the embedding vectors to have unit covariance. With simple algebraic formulation, this embedding cost problem can be transformed as follows: \begin{equation}\label{Lpcal Linear Preseving}
\begin{split}
&\mathop {\min }\limits_{\bm{Y}^v} \hspace{0.5em}tr\left( {\bm{Y}^v{{({\bm{I}_N} - \bm{W}^v)}^T}({\bm{I}_N} - \bm{W}^v){{\bm{Y}^v}^T}} \right) \\
&\hspace{0.5em} s.t. \hspace{1em}\bm{Y}^v{{\bm{Y}^v}^T} = {\bm{I}_d^v} \\
\end{split}
\end{equation} where $\bm{I}_N$ is a $N \times N$ unit matrix, $\bm{I}_d^v$ is a $d^v \times d^v$ unit matrix and $tr$ denotes the matrix trace.

It can be minimized by solving a sparse $N \times N$ eigenvalue problem of ${({\bm{I}_N} - \bm{W}^v)^T}({\bm{I}_N} - \bm{W}^v)$, whose bottom $d^v$ nonzero eigenvectors provide an ordered set of orthogonal coordinates centered on the origin.
\subsection{The construction of Multi-view Locality Low-rank Embedding for Dimension Reduction}\label{construction}
In this section, we introduce the the construction of $MvL^2E$ in detail. To integrate rich information among different features, the $L^2E$ for the $v$th view is extended into all views. By adding up cost function in Eq.(\ref{Lpcal Linear Preseving}) among all views, we could obtain the following optimization problem:
\begin{equation}
\begin{split}
&\mathop {\min }\limits_{{\bm{Y}^1},{\bm{Y}^2}, \ldots ,{\bm{Y}^m}} {\rm{ }} \sum\limits_{v = 1}^m {tr\left( {{\bm{Y}^v}{{(\bm{I} - {\bm{W}^v})}^T}(\bm{I} - {\bm{W}^v}){\bm{Y}^v}^{^T}} \right)} \\
& s.t. \hspace{0.5em}{\bm{Y}^v}{\bm{Y}^v}^{^T} = \bm{I},\forall 1 \le v \le m\ \\
\end{split}
\end{equation}

But this equation is equal to solve the $L^2E$ problem for all views separately and fails to integrate multi-view features to one common manifold space. For solving this existing problem, we propose a multi-view DR method called Multi-view Locality Low-rank Embedding for Dimension Reduction ($MvL^2E$) to fully apply all features from different views and learn a common low-dimensional representations. However, the dimension of the features set in each view owns its size, which is different from the other views. Besides, obtaining common manifold structure directly isn't easy to implement because of its intrinsic geometric properties in each view. Therefore, integrating different views into a common subspace is still full of challenges.

To address two issues above, we firstly make multi-view subspace hypotheses that the pairwise similarities of coefficient vectors are similar across all views. Then, we propose a novel embedding $\bm{Y}^*$ based on centroid to make $\bm{Y}^*$ that is closed to the low-dimensional embedding $\bm{Y}^v$ in the $v$th view. To deal with the dimensional difference problem between the centroid based embedding $\bm{Y}^*$ and the low-dimensional embedding $\bm{Y}^v$ in the $v$th view, we utilize the following cost function as a measurement of agreement  between the embedding $\bm{Y}^*$ of the centroid manifold structure and the embedding $\bm{Y}^v$ of the $v$th view:
\begin{equation}\label{F_norm}
S\left( {{\bm{Y}^*},{\bm{Y}^v}} \right) =  - \left\| {{\bm{K}^*} - {\bm{K}^v}} \right\|_F^2
\end{equation}
where $\bm{K}^*$ and $\bm{K}^v$ stand for the similarity matrix of the centroid $\bm{Y}^*$ and the $v$th view $\bm{Y}^v$ separately, $\left\|  \cdot  \right\|_F^2$ denotes the square of the Frobenius norm(F-norm) of the matrix. By utilizing the F-norms of the difference between the similarity $\bm{Y}^*$ and $\bm{Y}^v$ as measurement of agreement, it's convenient to solve the inconsistent dimension problem among all views. Although features from different views can reflect different properties of one sample, Eq.(\ref{F_norm}) guarantees that these features can share complementary information to help $MvL^2E$ to construct one common subspace. Obviously, the similarity matrix $\bm{K}^v$ for the $\bm{Y}^v$ has already taken care of the nonlinearities present in the $v$th view. Besides, using linear kernel usually could get a nice optimization problem. Hence, we choose the linear kernel for the $v$th view, i.e., $k(\bm{y}_i^v, \bm{y}_j^v)=\bm{y}_i^T\bm{y}_j$ as similar measurement in Eq.(\ref{F_norm}). This implies that $\bm{K}^v=\bm{Y}_v^T\bm{Y}_v$. So Eq.(\ref{F_norm}) could be expressed as follows:
\begin{equation}\label{F-norm2}
\begin{split}
&S\left( {{\bm{Y}^*},{\bm{Y}^v}} \right) = \left\| {{\bm{Y}^*}^{^T}{\bm{Y}^v} - {\bm{Y}^v}^{^T}{\bm{Y}^v}} \right\|_F^2 \\
&= 2tr\left( {{\bm{Y}^*}^{^T}{\bm{Y}^*}{\bm{Y}^v}^{^T}{\bm{Y}^v}} \right) - tr\left( {{\bm{Y}^*}^{^T}{\bm{Y}^*}{\bm{Y}^*}^{^T}{\bm{Y}^*}} \right) \\
&- tr\left( {{\bm{Y}^v}^{^T}{\bm{Y}^v}{\bm{Y}^v}^{^T}{\bm{Y}^v}} \right)    \\
\end{split}
\end{equation}

According to the constraint in the $L^2E$ loss function, it's easy to find that $tr\left( {{\bm{Y}^v}^{^T}{\bm{Y}^v}{\bm{Y}^v}^{^T}{\bm{Y}^v}} \right)$ is equal to a constant. Similarly, $tr\left( {{\bm{Y}^*}^{^T}{\bm{Y}^*}{\bm{Y}^*}^{^T}{\bm{Y}^*}} \right)$ is also equal to a constant. Substituting this into Eq.(\ref{F-norm2}) and ignoring the constants and scaling terms, we could get
\begin{equation}\label{agreement}
S\left( {{\bm{Y}^*},{\bm{Y}^v}} \right) = tr\left( {{\bm{Y}^*}^{^T}{\bm{Y}^*}{\bm{Y}^v}^{^T}{\bm{Y}^v}} \right)
\end{equation}

We maximize the agreement in Eq.(\ref{agreement}) to achieve the multi-view subspace hypotheses. Combining this with the $L^2E$ objectives of individual views, we can get the following maximization problem for $MvL^2E$:

\begin{equation}\label{$MvL^2E$}
\begin{split}
&\mathop {\max }\limits_{{\bm{Y}^*},{\bm{Y}^1},{\bm{Y}^2}, \ldots ,{\bm{Y}^m}} \gamma \sum\limits_{v = 1}^m {tr\left( {{\bm{Y}^*}^{^T}{\bm{Y}^*}{\bm{Y}^v}^{^T}\bm{Y}^v} \right)} \\
&- \sum\limits_{v = 1}^m {tr\left( {{\bm{Y}^v}{{(\bm{I} - {\bm{W}^v})}^T}(\bm{I} - {\bm{W}^v}){\bm{Y}^v}^{^T}} \right)} \\
& s.t.\hspace{0.5em}{\bm{Y}^*}{\bm{Y}^*}^{^T} = \bm{I},{\bm{Y}^v}{\bm{Y}^v}^{^T} = \bm{I},\forall 1 \le v \le m\ \\
\end{split}
\end{equation}
where $\gamma$ is a hyperparameter that controls the trade-off between the two terms of Eq.(\ref{$MvL^2E$}). The first term is the agreement between the centroid and all views to follow the multi-view subspace hypotheses. The second term is the $L^2E$ loss function from multiple views. For the features set from the $v$th view, its low-dimensional representations are $\bm{Y}= [\bm{y}_1^v,\bm{y}_2^v, \ldots ,\bm{y}_N^v]$. Through Eq.(\ref{$MvL^2E$}), we could find that different low-dimensional embedding $\bm{Y}^v$ inflect each other for the centroid representations. Therefore, the process of maximizing Eq.(\ref{$MvL^2E$}) aims to find a common subspace which can integrate features from multiple views and preserve local manifold structure as much as possible.

\subsection{Alternative Optimization}\label{optimization}
In this section, we derive the solution of $MvL^2E$ defined in Eq.(\ref{$MvL^2E$}), which is a nonlinearly constrained nonconvex optimization problem. To the best of our knowledge, there is no direct way to get a global optimal solution. For this reason, we propose an iterative alternating strategy based on the alternating optimization\cite{bezdek2002some} to obtain a local optimal solution.

First, we fix ${{\bm{Y}^1},{\bm{Y}^2}, \ldots ,{\bm{Y}^m}}$ to update $\bm{Y}^*$. The optimal problem in Eq.(\ref{$MvL^2E$}) is equivalent to the following optimization problem:

\begin{equation}
\begin{split}
    &\mathop {\max }\limits_{{\bm{Y}^*}} \gamma \sum\limits_{v = 1}^m {tr\left( {{\bm{Y}^*}^{^T}{\bm{Y}^*}{\bm{Y}^v}^{^T}{\bm{Y}^v}} \right)}\\
    & s.t.\hspace{0.5em}{\bm{Y}^*}{\bm{Y}^*}^{^T} = \bm{I}\\
\end{split}
\end{equation}

Due to the attributes of matrix trace, optimizing $\bm{Y}^*$ is equivalent to the following optimization problem:

\begin{equation}\label{solve_centroid}
    \begin{split}
        &\mathop {\max }\limits_{{\bm{Y}^*}} tr\left( {{\bm{Y}^*}\left( {\sum\limits_{v = 1}^m {\gamma {\bm{Y}^v}^{^T}{\bm{Y}^v}} } \right){\bm{Y}^*}^{^T}} \right)\\
        & s.t.\hspace{0.5em}{\bm{Y}^*}{\bm{Y}^*}^{^T} = \bm{I}\\
    \end{split}
\end{equation}

It's easy to find that $\bm{L}^*={\sum\limits_{v = 1}^m {\gamma {\bm{Y}^v}^{^T}{\bm{Y}^v}} }$ is symmetric. Based on the Ky-Fan theory\cite{bhatia2013matrix}, $\bm{Y}^*$ in Eq.(\ref{solve_centroid}) has a global optimal solution, which is given as the eigenvectors associated with the smallest $d^*$ eigenvalues of $\bm{L}^*={\sum\limits_{v = 1}^m {\gamma {\bm{Y}^v}^{^T}{\bm{Y}^v}} }$.

Second, we fix $\bm{Y}^*$ to update ${{\bm{Y}^1},{\bm{Y}^2}, \ldots ,{\bm{Y}^m}}$ separately. According to Eq.(\ref{$MvL^2E$}), it's not difficult to find that the optimal solution of the $v$th view is not depended on the other views when $\bm{Y}^*$ is fixed. Therefore, optimizing each view embedding $\bm{Y}^v$ can be expresses as the following problem:

\begin{equation}
    \begin{split}
        &\mathop {\max }\limits_{{\bm{Y}^*}} \gamma tr\left( {{\bm{Y}^*}^{^T}{\bm{Y}^*}{\bm{Y}^v}^{^T}\bm{Y}^v} \right)\\
        &- tr\left( {{\bm{Y}^v}{{(\bm{I} - {\bm{W}^v})}^T}(\bm{I} - {\bm{W}^v}){\bm{Y}^v}^{^T}} \right)\\
        & s.t. \hspace{0.5em}{\bm{Y}^v}{\bm{Y}^v}^{^T} = \bm{I}\\
    \end{split}
\end{equation}

According to the attributes of matrix trace, optimizing $\bm{Y}^v$ is equivalent to the following optimization problem:

\begin{equation}\label{solve_vth}
    \begin{split}
        &\mathop {\max }\limits_{{\bm{Y}^*}} tr\left( {{\bm{Y}^v}\left( {\gamma{\bm{Y}^*}^{^T}{\bm{Y}^*}
        - {{(\bm{I} - {\bm{W}^v})}^T}(\bm{I} - {\bm{W}^v})} \right){\bm{Y}^v}^{^T}} \right)\\
        &\hspace{0.5em} s.t. \hspace{0.5em}{\bm{Y}^v}{\bm{Y}^v}^{^T} = \bm{I}\\
    \end{split}
\end{equation}

Because both ${\bm{Y}^*}^{^T}{\bm{Y}^*}$ and ${{(\bm{I} - {\bm{W}^v})}^T}(\bm{I} - {\bm{W}^v})$ are symmetric, it can be easily inferred that $\gamma {{\bm{Y}^*}^{^T}{\bm{Y}^*}- {{(\bm{I} - {\bm{W}^v})}^T}(\bm{I} - {\bm{W}^v})} $ is also symmetric. Based on the Ky-Fan theory\cite{bhatia2013matrix}, $\bm{Y}^v$ in Eq.(\ref{solve_vth}) has a global optimal solution, which is given as the eigenvectors associated with the smallest $d^v$ eigenvalues of  $\gamma {\bm{Y}^*}^{T}{\bm{Y}^*}$ -  ${{(\bm{I} - {\bm{W}^v})}^T}(\bm{I} - {\bm{W}^v}) $. For all views, ${{\bm{Y}^1},{\bm{Y}^2}, \ldots ,{\bm{Y}^m}}$ could be solved separately by the optimization strategy above.

According to the descriptions above, we can form an alternating optimization strategy, summarized in \textbf{Algorithm 1},  to capture a local optimal solution of $MvL^2E$.


\begin{algorithm}
\caption{The optimization procedure of $MvL^2E$}
\hspace*{0.02in} {\bf Input:}

\hspace*{0.05in}1. A multi-view features set with N training samples having m views ${X^v} = [x_1^v,x_2^v, \ldots ,x_N^v] \in {\mathbb{R}^{{D_v} \times N}}$.

\hspace*{0.05in}2. The regularization parameter $\gamma$ in Eq.(\ref{$MvL^2E$}).

\hspace*{0.02in} {\bf Output:}\hspace*{0.08in}The centroid embedding $Y^*$

\hspace*{0.02in} {\bf The Main Procedure:}
\begin{algorithmic}
\FOR{v=1:m}
    \STATE{3. Obtain $W^v$ for the $v$th view by solving Eq.(\ref{$L^2E$}) and filling transform.}
    \STATE{4. Initialize $Y^v$ for the $v$th view by solving Eq.(\ref{Lpcal Linear Preseving}) using the eigenvalue decomposition method.}
\ENDFOR
\REPEAT
\STATE{5. Update $Y^*$ by solving Eq.(\ref{solve_centroid}).}
\FOR{v=1:m}
    \STATE{6. Update $Y^v$ for the $v$th view by solving Eq.(\ref{solve_vth}).}
\ENDFOR

\UNTIL{$Y^*$ converges}
\end{algorithmic}

\end{algorithm}

Texts and images are usually represented by multi-view features, and the feature in each view is represented in high-dimensional space. In this section, we evaluate the performance of $MvL^2E$ by comparing with several classical DR methods and multi-view learning methods in the multi-view datasets of texts and images. These experiments results verify the excellent performance of $MvL^2E$.
\subsection{Datasets and Comparing Methods}
There are five datasets in form of texts and images. Two text datasets adopted in the experiments are widely used in works, including 3Source\footnotetext[1]{http://mlg.ucd.ie/datasets/3sources.html}, Cora\footnotetext[2]{3http://lig-membres.imag.fr/grimal/data.html}. 3Sources consist of 3 well-known online news sources: BBC, Reuters and the Guardian, and each source is treated as one view. We select the 169 stories which are reported in all these 3 sources; Cora consists of 2708 scientific publications which come from 7 classes. Because document is represented by content and cites views, Cora could be considered as a two views datasets. Three images datasets adopted in the experiments are widely used in works, including: ORL\footnotetext[3]{http://www.uk.research.att.com/facedatabase.html}, Yale\footnotetext[4]{http://cvc.yale.edu/projects/yalefaces/yalefaces.html}, Caltech 101\footnotetext[5]{http://www.vision.caltech.edu/ImageDatasets/Caltech101/}.

\begin{figure}[!htb]
\centering
\includegraphics[width=5in]{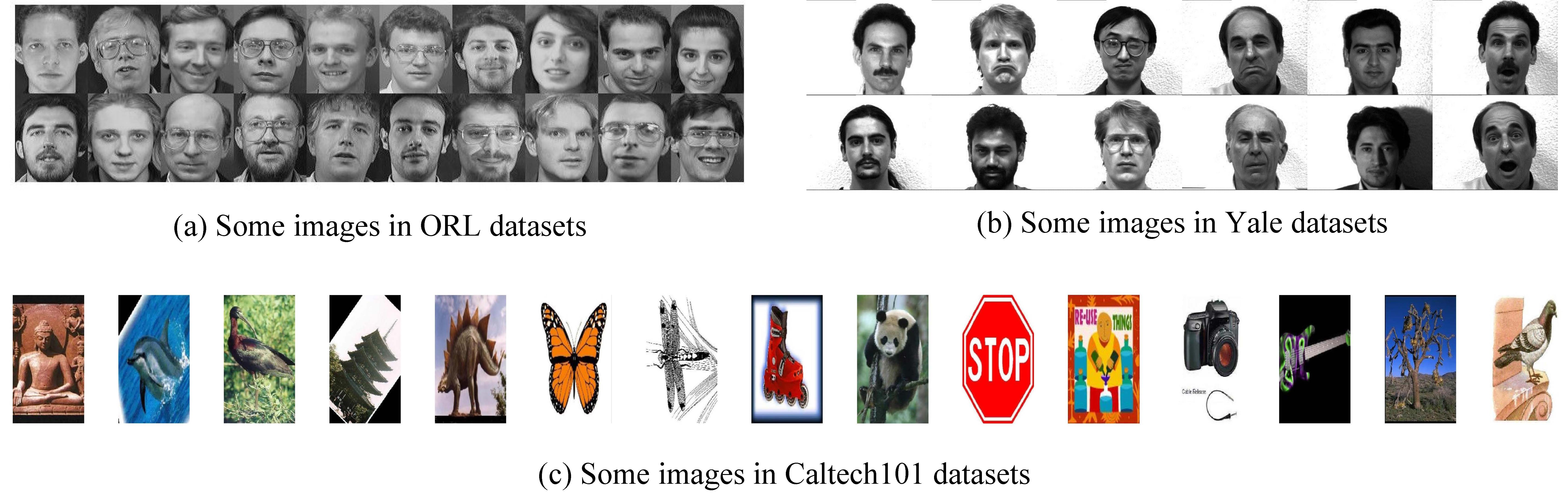}
\caption{Examples Images}
\label{Fig2}
\end{figure}

ORL and Yale are two face image datasets which have been widely used in face recognition. Caltech101 is a benchmark image dataset which contains 9144 images corresponding to 102 objects. We extract features for images using three different image descriptors. The detailed information of these datasets is summarized in table \ref{table1}. Some example images in image datasets are shown in the Fig.\ref{Fig2}.

\begin{table}[!htb]
\caption{The detail information of the multi-view datasets}
\centering
\begin{tabular}{lccc} 
\hline
Datasets &Samples &Classes &Views\\
\hline  
Sources &169 &6 &3\\
Cora &2708 &7 &2\\
ORL &400 &40 &3\\
Yale &165 &15 &3\\
Caltech101 &9144 &102 &3\\
\hline
\end{tabular}
\label{table1}
\end{table}

\section{Experiments}\label{Related Materials}

The effectiveness of $MvL^2E$ is evaluated by comparing the following algorithms, including: the best performance of the single view based LLE(BLLE), the best performance of the single view based LE(BLE), the feature concatenation based LLE(CLLE), MSE, Co-regularized, CCA. Besides, we also compared the single view low-dimensional embedding in our framework with original low-dimensional embedding using $L^2E$, and additional experiments on the single feature in multi-view framework by correcting and complemented by ones from the others views are to verify the fact that our method could make use of complementary information among different views by correcting and complementing ones from the others views.

\subsection{Parameter Setting and Evaluation methods}
In our experiments, we set the hyperparameter $\gamma=0.8$ in Eq.(\ref{$MvL^2E$}). And we will demonstrate the reason that we choose $\gamma=0.8$ in Section \ref{hyperparameter influence}. All DR methods are evaluated 30 times with different random training samples and testing samples, and the mean(MEAN) and max(MAX) classification accuracies on multi-view datasets are employed as the evaluation index.

\subsection{Experiments on textual datasets}
In an attempt to show the superior performance of $MvL^2E$, the experiments on two multi-view textual datasets (3Source, Cora) are shown in this section. And 1NN classifier is adopted here to classify all testing samples to verify the performances of all DR methods when we have obtained the low-dimensional embedding using all DR methods.

For 3Source dataset, we randomly select 80\% of the samples for each subset as training samples every times. The dimension of embedding obtained by all DR methods all maintains 30 dimensions. We run all DR methods 30 times with different random training samples and testing samples. Table \ref{table2} shows the MEAN and MAX value on 3Source dataset.

\begin{table}[!htb]
\caption{The classification accuracy on 3Source dataset}
\label{table2}
\centering
\begin{tabular}{lllll}
\hline
DR Methods & \multicolumn{2}{c}{Dims=20} & \multicolumn{2}{c}{Dims=30} \\
 & MEAN(\%) & MAX(\%) & MEAN(\%) & MAX(\%)\\
\hline
BLLE & 69.9 & 79.1 & 72.7 & 79.8 \\
BLE & 71.6 & 75.4 & 68.7 & 75.8 \\
CLLE & 77.3 & 88.2 & 78.3 & 85.2 \\
MSE & 79.3 & 90.5 & 79.8 & 91.0 \\
Co-regularized & 79.5 & 89.1 & \textbf{82.4} & 90.5  \\
CCA  & 53.8 & 76.4 & 54.7 & 73.5 \\
$MvL^2E$ & \textbf{82.7} & \textbf{90.5} & 81.7 & \textbf{91.9} \\
\hline
\end{tabular}
\end{table}

For Cora dataset, we randomly select 80\% of the samples for each subsets as training samples every times. The dimension of embedding obtained by all DR methods all maintains 20 dimensions and 30 dimensions. We run all DR methods 30 times with different random training samples and testing samples. Table \ref{table3} shows the MEAN and MAX value on Cora dataset.

Through tables \ref{table2}-\ref{table3}, we can clearly find that $MvL^2E$ outperforms the other 6 DR methods in most situations. And CLLE that concatenates features from different views couldn't gain a good performance. Therefore, our framework for multi-view features are more effective. Because $MvL^2E$ can integrate compatible and complementary information from multi-view features, $MvL^2E$ can obtain a more excellent performance.

\begin{table}[htbp]
\caption{The classification accuracy on Cora dataset}
\label{table3}
\centering
\begin{tabular}{lllll}  
\hline
DR Methods & \multicolumn{2}{c}{Dims=20} & \multicolumn{2}{c}{Dims=30} \\
 & MEAN(\%) & MAX(\%) & MEAN(\%) & MAX(\%)\\
\hline
BLLE & 61.3 & 65.6 & 60.9 & 66.7 \\
BLE & 61.7 & 66.3 & 64.7 & 68.5 \\
CLLE & 46.3 & 49.9 & 54.5 & 58.3 \\
MSE & 40.3 & 42.8 & 40.7 & 44.6 \\
Co-regularized & 60.6 & 63.6 & 60.0 & 62.3 \\
CCA  & 71.1 & 73.8 & 71.5 & 74.3 \\
$MvL^2E$ & \textbf{73.8} & \textbf{75.4} & \textbf{74.1} & \textbf{76.8} \\
\hline
\end{tabular}
\end{table}

\subsection{Experiments on images datasets}
In an attempt to show the superior performance of $MvL^2E$, the experiments on three multi-view images datasets (Yale, ORL, Caltech101) are shown in this section. And 1NN classifier is adopted here to classify all testing samples to verify the performances of all DR methods when we have obtained the low-dimensional embdedings using all DR methods.

For Yale dataset,we extract gray-scale intensity, local binary patterns and edge direction histogram as 3 views. The dimension of embedding obtained by all DR methods all maintains 20 dimensions and 30 dimensions. We randomly select 80\% of the samples for each subsets as training samples every times and run all DR methods 30 times with different random training samples and testing samples. Table \ref{table4} shows the MEAN and MAX value on Yale dataset.

\begin{table}[htbp]
\caption{The classification accuracy on Yale dataset}
\label{table4}
\centering
\begin{tabular}{lllll}
\hline
DR Methods & \multicolumn{2}{c}{Dims=20} & \multicolumn{2}{c}{Dims=30} \\
 & MEAN(\%) & MAX(\%) & MEAN(\%) & MAX(\%)\\
\hline
BLLE &67.4 & 71.5 & 87.5 & 90.2 \\
BLE & 81.2 & 85.4 & 80.1 & 83.4 \\
CLLE & 64.6 & 78.7 & 64.8 & 75.7 \\
MSE & 74.4 & 85.1 & 61.6 & 92.9 \\
Co-regularized &76.4 & 90.9 & 69.3 & 84.8 \\
CCA   & 80.7 & \textbf{90.8} & 81.8 &  91.6\\
$MvL^2E$ & \textbf{80.9} & 88.0 & \textbf{89.6} & \textbf{96.0}  \\
\hline
\end{tabular}
\end{table}

\begin{figure}[!htb]
\centering
\includegraphics[width=4in]{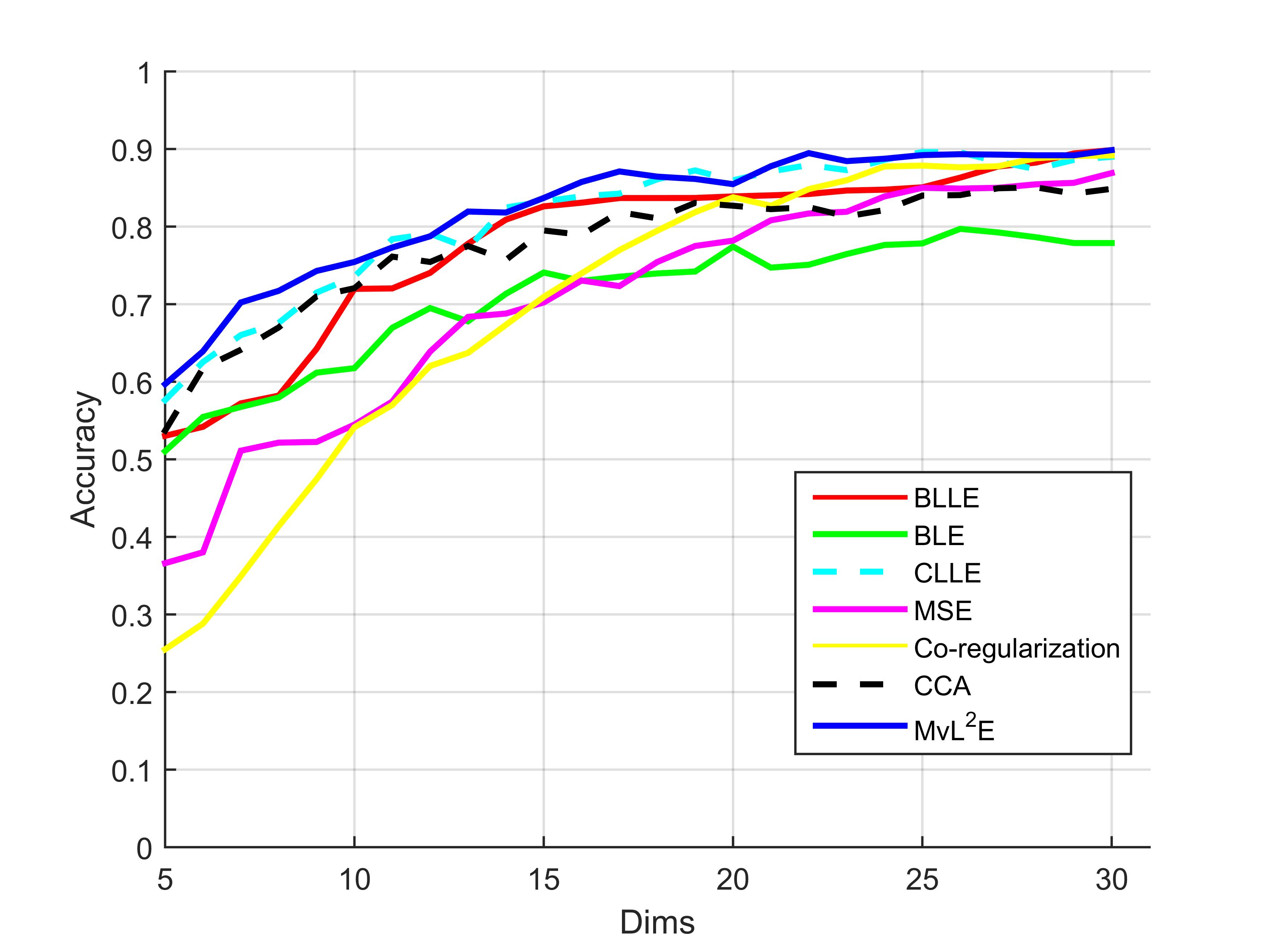}
\caption{The classification accuracy on ORL dataset}
\label{Fig3}
\end{figure}

For ORL dataset, we extract gray-scale intensity, local binary patterns and edge direction histogram as 3 views. The dimension of embedding obtained by all DR methods all maintains from 5 to 30 dimensions. We randomly select 80\% of the samples for each subsets as training samples every times and run all DR methods 30 times with different random training samples and testing samples. Fig.\ref{Fig3} shows the mean accuracy values on ORL dataset.

\begin{figure}[!htb]
\centering
\includegraphics[width=4in]{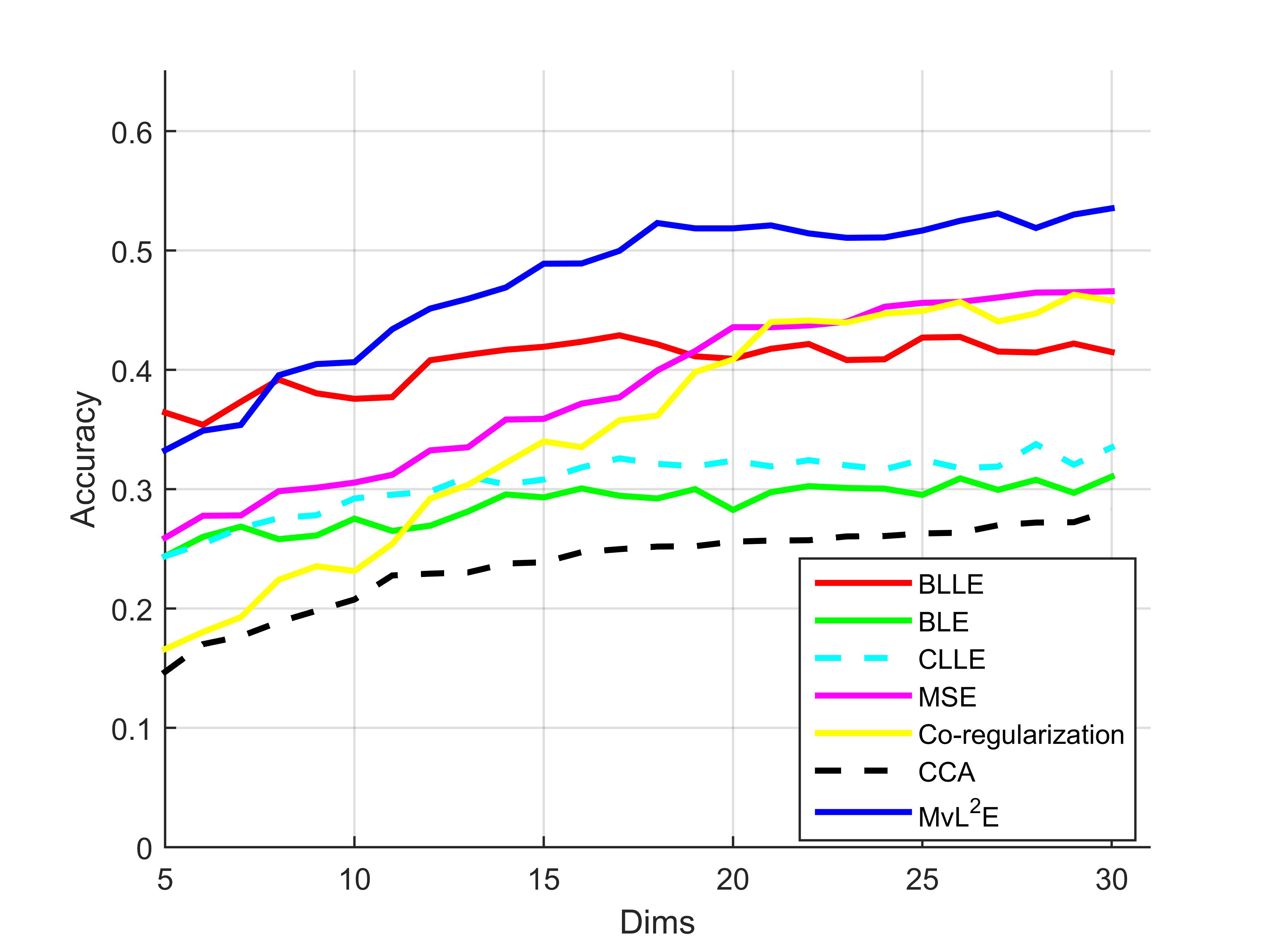}
\caption{The classification accuracy on Caltech101 dataset}
\label{Fig4}
\end{figure}

For Caltech101 dataset, the first 20 classes are utilized in our experiments. Meanwhile, we extract gist, local binary patterns and edge direction histogram as 3 views. The dimension of embedding obtained by all DR methods maintains from 5 to 30 dimensions. We randomly select 80\% of the samples for each subsets as training samples every times and run all DR methods 30 times with different random training samples and testing samples. Fig.\ref{Fig4} shows the mean accuracy values on Caltech101 dataset.

Through table \ref{table4}, Fig.\ref{Fig3}-\ref{Fig4}, we can clearly find that $MvL^2E$ outperforms the other 6 DR methods in most situations. And CLLE that concatenating features from different views couldn't gain a good performance. Therefore, our framework for multi-view features are more effective. In summary, $MvL^2E$ could integrate compatible and complementary information from multi-view features and obtain a more excellent performance.

\subsection{Comparison between Single View Embedding}

Besides the effectiveness of the centroid manifold embedding, we also find that the single feature in our multi-view framework obtains more outstanding performance than original manifold space by correcting and complemented by ones from the others views. To verify this opinion above, we evaluate the performance of $L^2E$ of single view in our framework by comparing with original $L^2E$ of single view in five datasets, including 3Source, Cora, Yale, ORL and Caltech101. For all datasets above, we choose the second view as compared view and the dimension of embedding maintains 20 dimensions. We run this two methods among all datasets 30 times with different random training samples and testing samples. Table \ref{table5} shows the MEAN and MAX value on all datasets.

\begin{table}[!htb]
\caption{The classification accuracies of different $L^2E$ methods}
\label{table5}
\centering
\begin{tabular}{lllll}  
\hline
DATASETS & \multicolumn{2}{c}{$L^2E$ in our framework} & \multicolumn{2}{c}{Original $L^2E$} \\
 & MEAN(\%) & MAX(\%) & MEAN(\%) & MAX(\%)\\
\hline
3Source & 82.4 & 92.2 & 76.1 & 90.2 \\
Cora & 75.7 & 77.6 & 60.4 & 64.3 \\
Yale & 68.6 & 78.0 & 44.8 & 53.6 \\
OLR & 82.4 & 91.6 & 68.8 & 77.5 \\
Caltech101 & 43.6 & 49.2 & 30.8 & 35.3 \\
\hline
\end{tabular}
\end{table}

Through table \ref{table5}, we could find that the single feature in our multi-view framework obtains more outstanding performance than original manifold. Therefore, the $L^2E$ of single view in our framework is also more effective. Not only the low-dimensional feature in a common manifold space has reliable performance, but also the single view feature obtains more outstanding performance than original manifold space by correcting and complemented by ones from the others views.

\subsection{Convergence of $MvL^2E$}

\begin{figure}[!htb]
\centering
\includegraphics[width=5in]{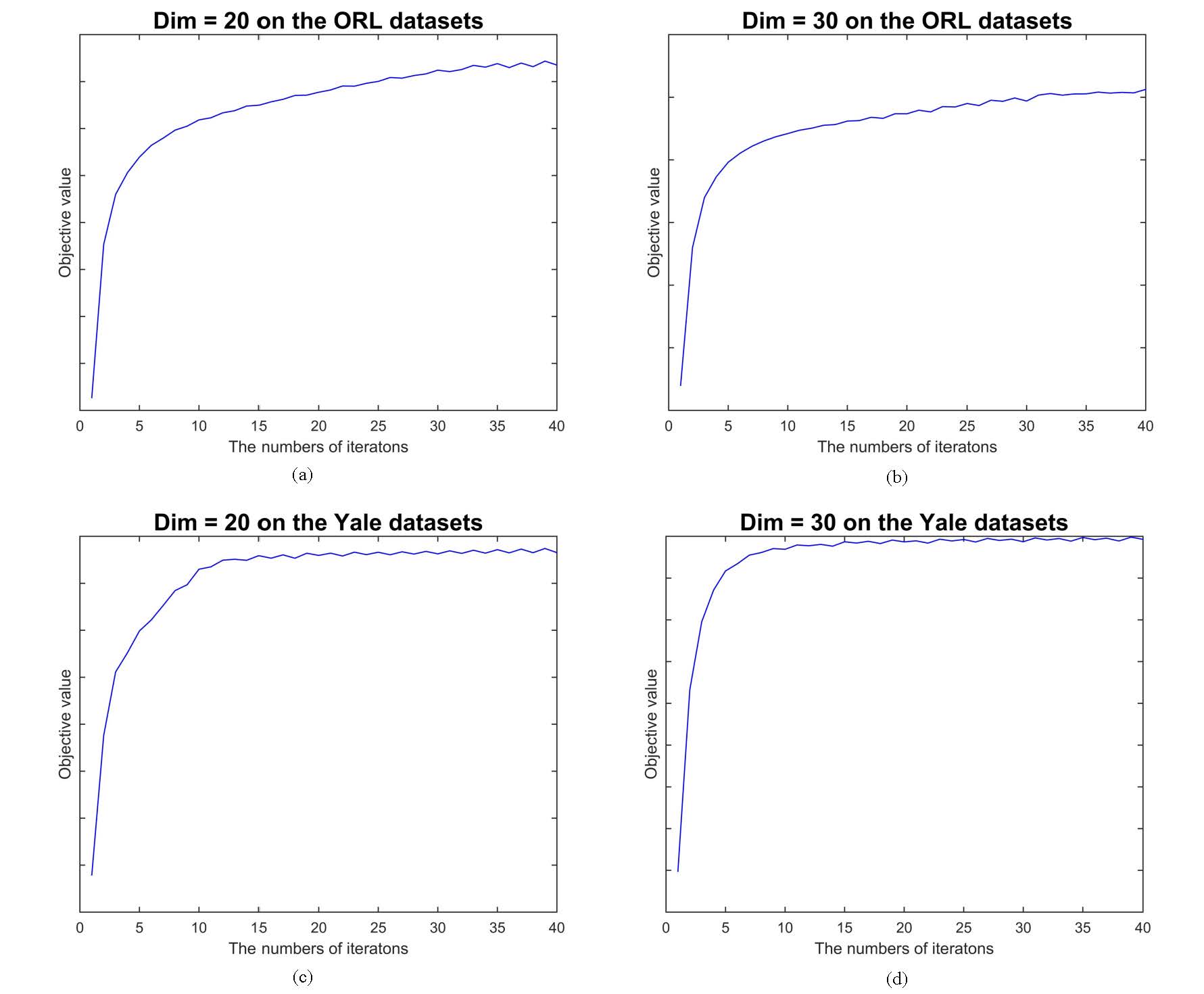}
\caption{Objective values on ORL and Yale}
\label{Fig5}
\end{figure}

Because $MvL^2E$ adopts an iterative procedure to obtain the optimal solution, it is essential to discuss the convergence and training time of $MvL^2E$ in detail. In this section, we summarize the objective values on ORL and Yale datasets according to the experiments above.All the training parameters (such as training numbers, dimensions) can be found above Fig.\ref{Fig5}, which summarizes the objective values of ORL and Yale datasets.

We can clearly find in Fig.\ref{Fig5} that the curve of the objective values tends to be stable after ten iterations on the Yale datasets and the objective values tend to be stable after thirty iterations on the ORL datasets. It implies the fact that $MvL^2E$ could converge within a limited number of iterations, and the size of the matrix is an important factor affecting the speed of convergence according to the different iterations numbers of the ORL and Yale datasets. Besides, the dimensionality of embedding also has some impact, and the training time increases when the dimensionality of the embedding raises.

\subsection{Analysis of the hyperparameter influence}\label{hyperparameter influence}

\begin{figure}[!htb]
\centering
\includegraphics[width=5in]{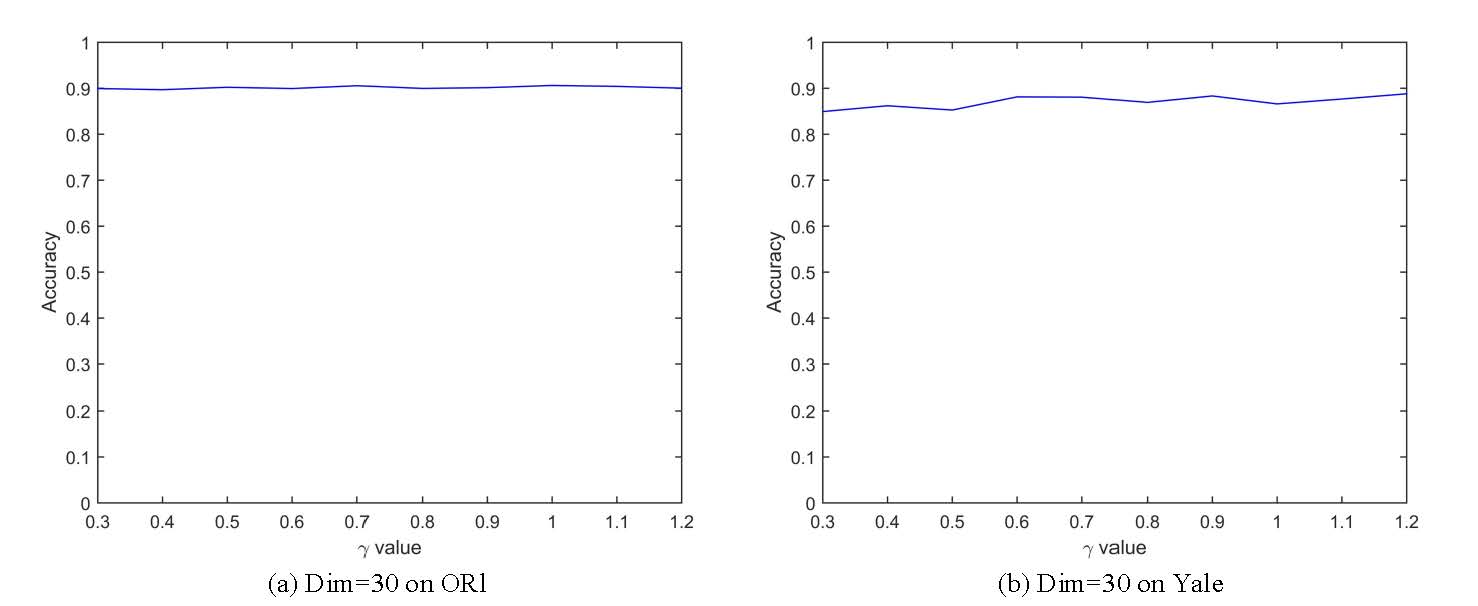}
\caption{Objective values on ORL and Yale}
\label{Fig6}
\end{figure}

To fully validate the effectiveness of $MvL^2E$, this subsection mainly analyzes the influences on the performance of the parameter $\gamma$ introduced in $MvL^2E$, where $\gamma$ is employed as trade-off parameter to balance the multi-view agreement term and $L^2E$ loss term. As is shown in Fig.\ref{Fig6}, which summarizes the classify  accuracy values of ORL and Yale datasets, where the dimensionality of low-dimensional emdedding is 30. Even the performance increases with the increase in $\gamma$, the oscillation of accuracy becomes very stable in general. Especially, the accuracy will tend to a stable fixed point when $\gamma$ grows more than 0.8. Therefore, we set the hyperparameter $\gamma=0.8$ in Eq.(\ref{$MvL^2E$}).

\section{Conclusion}
Prior work has documented the effectiveness of traditional DR methods on processing high-dimensional features and reducing the time consumption and computation cost, such as PCA, LDA, and etc. However, these studies have either been not directly extended into the multi-view framework or have not focused on compatibility and complementary among different views. In this study, we investigate the low rank linear local structure in the $v$th view and try to extend it to directly process multi-view features. We find that local structure with low rank property has nice robustness and projecting all views into a common space is feasible. Therefore, in this paper, we first propose a new DR method called Locality Low-rank Embedding ($L^2E$) which maintains the low rank local linear structure in the geometric manifold space. Then we propose a multi-view method called Multi-view Locality Low-rank Embedding for Dimension Reduction ($MvL^2E$) extending the $L^2E$ for the single view to the multi-view framework, which fully integrates compatible and complementary information from multi-view features sets to construct low-dimensional embedding. Our results provide compelling evidence that  $MvL^2E$ is an effective multi-view DR method and suggest that the single feature in each view obtains more outstanding performance than original manifold space by correcting and complemented by ones from the others views. However, one limitation is worth noting that solving $MvL^2E$ needs to perform eigenspace decomposition of the matrix of $N \times N$ size ($N$ is the number of samples in the dataset). This generally takes O($N^3$) time, and it will be very time consuming when N is very large. In the future, we will consider how to utilize the sampling technique\cite{talwalkar2008large} in $MvL^2E$ to handle a large-scale dataset.

\section*{Acknowledgment}
The authors would like to thank the anonymous reviewers for their insightful comments and the suggestions to significantly improve the quality of this paper. This work was supported by National Natural Science Foundation of PR China(61672130) and LiaoNing Revitalization Talents Program(XLYC1806006).

\bibliography{mybibfile}

\indent  

\begin{window}[0,l,{\mbox{\includegraphics[width=1in]{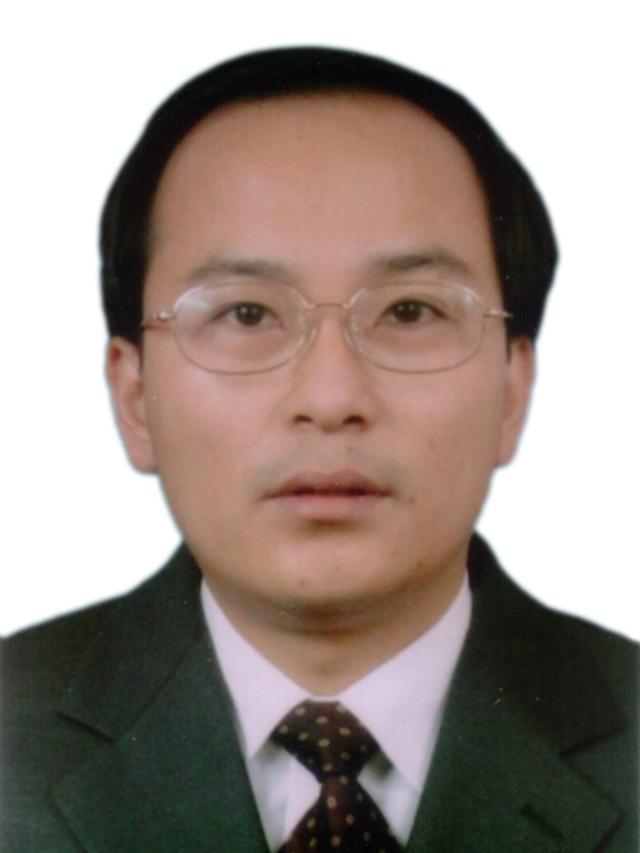}}},{}]
\textbf{Lin Feng} received the BS degree in electronic technology from Dalian University of Technology, China, in 1992, the MS degree in power engineering from Dalian University of Technology, China, in 1995,and the PhD degree in mechanical design and theory from Dalian University of Technology, China, in 2004. He is currently a professor and doctoral supervisor in the School of Innovation Experiment, Dalian University of Technology, China. His research interests include intelligent image processing, robotics, data mining, and embedded systems.
\end{window}
\indent  
\begin{window}[0,l,{\mbox{\includegraphics[width=1in]{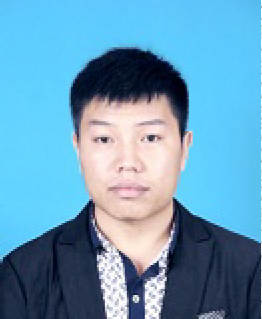}}},{}]
\textbf{Xiangzhu Meng} received his BS degree from Anhui University, in 2015. Now he is working towards the PHD degree in School of Computer Science and Technology, Dalian University of Technology, China. His research interests include mulit-view learning, deep learning and computing vision.
\end{window}
\indent  
\begin{window}[0,l,{\mbox{\includegraphics[width=1in]{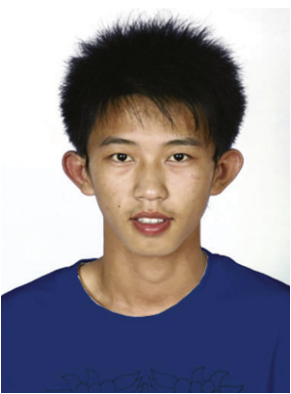}}},{}]
\textbf{Huibing Wang} received the Ph.D. degree in the School of Computer Science and Technology, Dalian University of Technology, Dalian, in 2018. During 2016 and 2017, he is a visiting scholar at the University of Adelaide, Adelaide, Australia. Now, he is a postdoctor in Dalian Maritime University, Dalian, Liaoning, China. He has authored and co-authored more than 20 papers in some famous journals or conferences, including TMM, TITS, TSMCS, ECCV, etc.
Furthermore, he serves as reviewers for TNNLS, Nurocomputing, PR Letters and MTAP, etc. His research interests include computing vision and machine learning
\end{window}

\end{document}